\documentclass{llncs}

\usepackage{graphicx}
\usepackage{multirow}
\usepackage{listings}
\usepackage{amsmath}  % for equation*
\usepackage{array}    % for tabular
\usepackage{verbatim} % for comment
\usepackage{wrapfig}
\usepackage{tabularx}
\usepackage{amssymb}
\usepackage{xcolor}
\usepackage{microtype}
\usepackage{url}
\usepackage{framed}
\usepackage{cleveref}
\usepackage{todonotes}
\usepackage{float}
\usepackage{caption}

\begin{document}

\title{Identifying cross country skiing techniques using power meters in ski poles} 
\author{Moa Johansson\inst{1}, Marie Korneliusson\inst{1}, Nickey Lizbat Lawrence\inst{1}}
\institute{Chalmers University of Technology, Gothenburg, Sweden. \\
	\email{moa.johansson@chalmers.se, \{marieko, nickey\}@student.chalmers.se}
	}
\maketitle

\begin{abstract}
Power meters are widely used for measuring training and racing effort in cycling, and the use of such sensors is now spreading also to other sports. 
Increasing volumes of data can be collected from athletes, with the aim of helping coaches and athletes analyse and understand training load, racing efforts, technique etc.
In this project, we have collaborated with Skisens AB, a company producing handles for cross country ski poles equipped with power meters. We have conducted a pilot study on the use of machine learning techniques on data from Skisens poles to identify which sub-technique a skier is using (double poling or gears 2-4 in skating), based only on the sensor data from the ski poles. The dataset contained labelled time-series data from three individual skiers using four different gears recorded in varied locations and varied terrain. 
We systematically evaluated a number of machine learning techniques based on neural networks with best results obtained by a LSTM network (accuracy of 95\% correctly classified strokes), when a subset of data from all three skiers was used for training. As expected, accuracy dropped to 78\% when the model was trained on data from only two skiers and tested on the third. To achieve better generalisation to individuals not appearing in the training set more data is required, which is ongoing work.

\end{abstract}

\section{Introduction}\label{sec:intro}
In a professional cross country ski race, as in many other sports, the first thing the athletes do after crossing the finish line is often to switch off their smart sports-watch. Why? 

The development of a wide range of sensors and products such as GPS-sensors, heart-rate monitors, motion sensors and power sensors have made it possible to record a vast amount of data from athletes, providing a rich source of information to help coaches and athletes measure, analyse and understand training load, racing efforts and technique. Sports like cycling has lead the way among the endurance sports, as it its relatively easy to equip a bicycle with various sensors, for instance, to accurately measure the power in each pedal stroke. Using power meters to steer training effort has become common not only for professional cyclists and coaches, but also for more ambitions recreational riders \cite{allen-coggan}. Given the relative ease at which large volumes of data can be recorded from sensors, we believe that machine learning has the potential to provide valuable tools for assisting data analysis in sports. 

In this project, we have collaborated with Skisens AB, a spin-off company from Chalmers University of Technology, which produces a power meter for cross-country skiing, mounted inside the handle of the pole. Unlike cycling where all power comes from the legs via the pedals, in skiing the proportion of power measured in the poles depends on skiing technique.
Broadly speaking, the skiing techniques may be divided into classical style and freestyle, each regulated by rules in competition. Furthermore, the two styles can each be broken down into several sub-techniques. The most effective sub-technique will depend on the terrain, the snow conditions and the
individual strengths of the skier (we give a brief introduction to cross-country skiing techniques in section  \ref{sec:background}). In order for an athlete and/or coach to accurately analyse the effort based on data recorded from a race it is therefore valuable to be able to get an automated classification of which technique was used where during the race. This work focuses on free-style technique, however, the methods may be applied also to classical style.

% the power produced from the arms through the poles depends on which \emph{technique} (or gear) the skier is using, broadly falling into two groups, classical and freestyle techniques. 
%The most effective technique will depend on the terrain, the snow conditions and the individual strengths of the skier (we give a brief introduction to cross-country skiing techniques in section \ref{sec:background}).   
%In order for an athlete and/or coach to accurately analyse the effort based on data recorded from a race it is therefore valuable to be able to get an automated classification of which technique was used where during the race. 

We use a dataset provided by Skisens, containing data from three skiers using Skisens handles while roller-skiing using different techniques in varied terrain. The dataset and data pre-processing is described in section \ref{sec:dataset}. We have evaluated three frequently used kinds of \emph{deep neural network} classifiers on this dataset: A convolutional neural network (CNN), a Long-Short Term Memory (LSTM) network \cite{LSTM}, and finally a bi-directional LSTM (BLSTM) model \cite{BLSTM}, described in more detail in section \ref{sec:models}. 
The set up of our study is inspired by Hammerla et al. \cite{hammerla2016deep}, who experimented thoroughly with these kinds of deep neural network to classify a variety of human movements using data from wearable sensors (e.g. household activities, physical exercise as well as gait abnormalities arising in Parkinson's disease). 
We have experimentally evaluated the models in two experiments (see section \ref{sec:experiments}): the first used a subset of data from all skiers for training in which the LSTM model reached the best accuracy (95\% on unseen test data), and a second experiment where this model was trained on data from two skiers and evaluated on the third. As expected, the accuracy for the LSTM model then dropped to 78\% on unseen test data.

There has been several previous works aiming at classifying cross-country skiing technique using a variety of sensors. Marshland et. al equipped cross-country skiers with a sensor unit attached on the skiers back, and observed that there were sufficient regularities in the sensor data which would motivate the development of algorithmic techniques for technique identification \cite{marshland}. This has been followed by several studies using different combinations of sensors and machine learning techniques, with promising results. St{\"o}ggl et al. used accelerometer data from a mobile phone attached to a belt around the chest of the skier and a Markov chain model to classify strokes \cite{holst2013,Stoggl2014}. When trained and tested on the same individuals, their algorithm reached an accuracy of 90.3\% $\pm$  4.1\%, which dropped to 86.0\% $\pm$ 8.9\% when trained on collective data. 
Rindal et al. used wearable inertial measurement units (IMUs) attached to the skiers arms and chest, together with gyroscopes attached to the skiers arms, to classify classical skiing techniques \cite{Rindal2017}. The gyroscopes helped identifying each stroke cycle, and the IMU-data was used to train a neural network classifier reaching an accuracy of 93.9\%. Sakurai et al. also used data from several IMUs attached to the skis and poles to construct a decision tree classifier both for classical and skating techniques \cite{sakurai2014,sakurai2016}.  Recently, Jang et al. conducted a study using wearable gyroscope sensors to identify both classical and skating techniques and a deep machine learning model combining CNN and LSTM layers \cite{Jang2018}. The best results were obtained with sensors attached to both hands, both feet and the pelvis, which reached an accuracy of 80\% when two skiers were used for training, and an unseen for testing, rising to between 87.2\% to 95.1\% (depending on terrain) when three skiers were used for training, and a forth unseen one for testing.

The main difference between our work and the above ski technique classifiers is that we do not use any dedicated wearable sensors for the task, but simply explore if we can identify technique using only the sensors already present in the Skisens pole for measuring power. Our sensor data only records the movements of the hands, and does not include any sensors on the body or on the skis, which would make the task easier. Nevertheless, we reach comparable or better accuracy results. Another advantage of using deep neural networks is that they do not require hand crafted features to be passed to the model. %by using deep neural network models, something which has not been much explored in other studies and has the advantage of not requiring hand crafted features to be passed to the model. 

\section{Background: Cross country skiing techniques}
\label{sec:background}
%We should briefly explain what the different gears are, perhaps including some pictures.

\begin{figure}[htbp]
\includegraphics[width=0.32\textwidth]{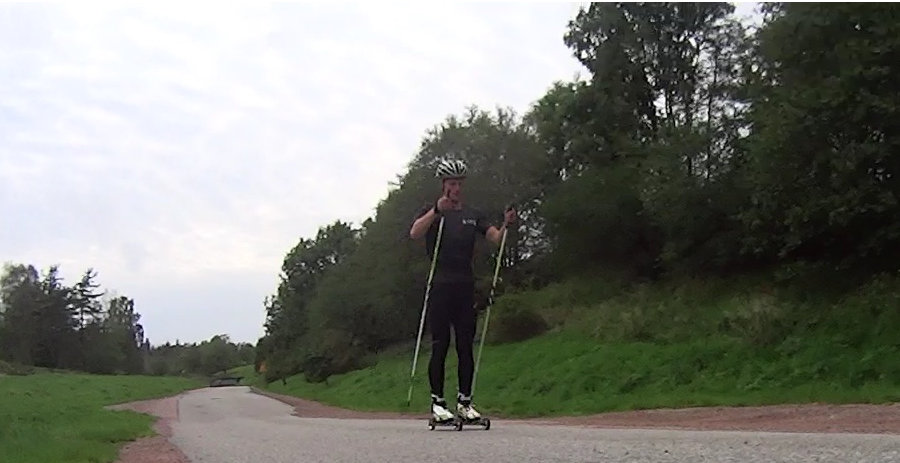}
\includegraphics[width=0.32\textwidth]{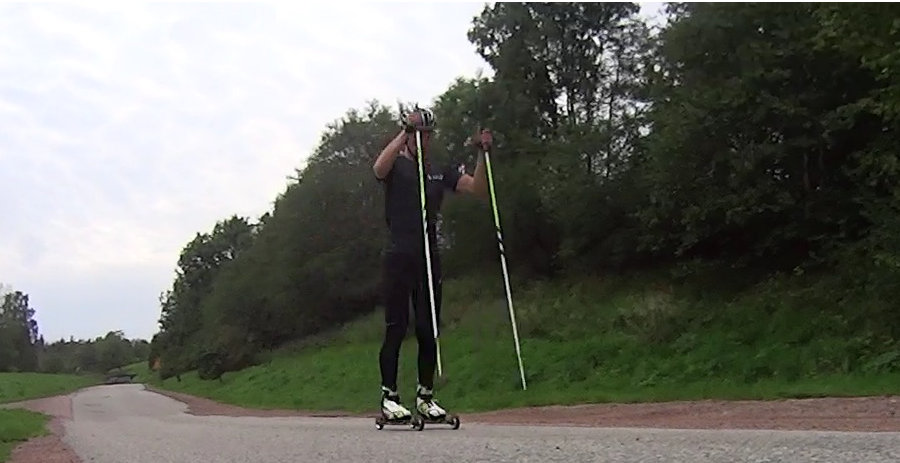}
\includegraphics[width=0.32\textwidth]{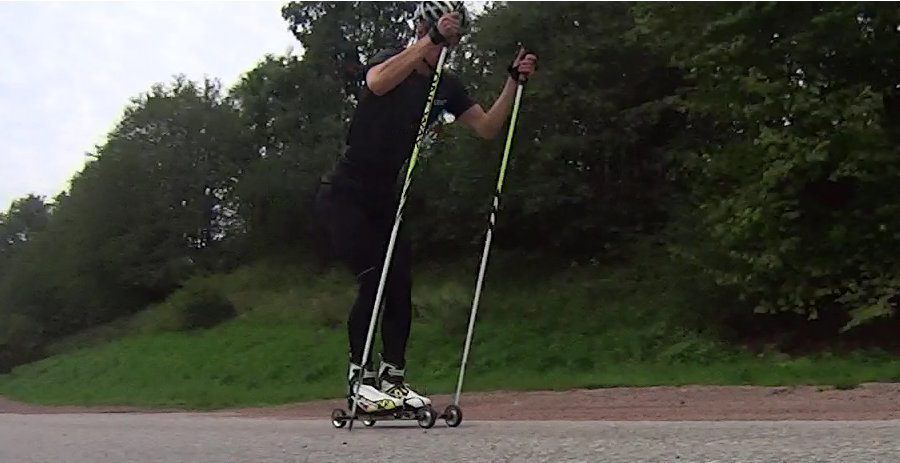}

\includegraphics[width=0.32\textwidth]{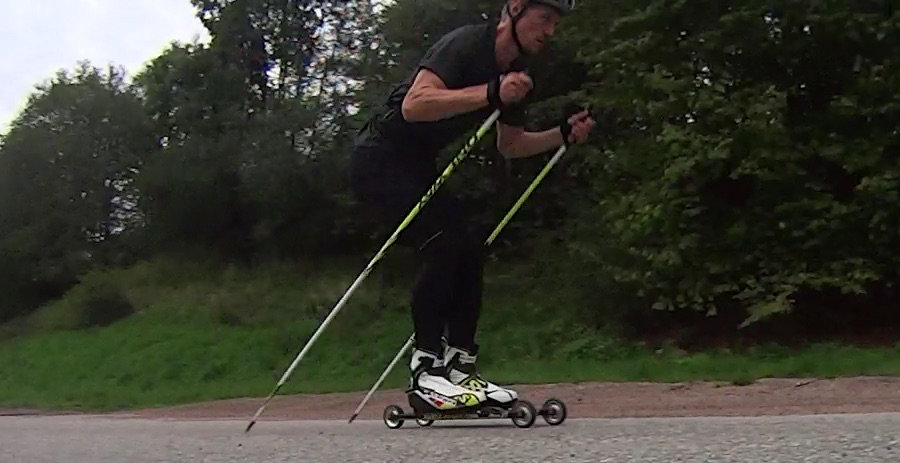}
\includegraphics[width=0.32\textwidth]{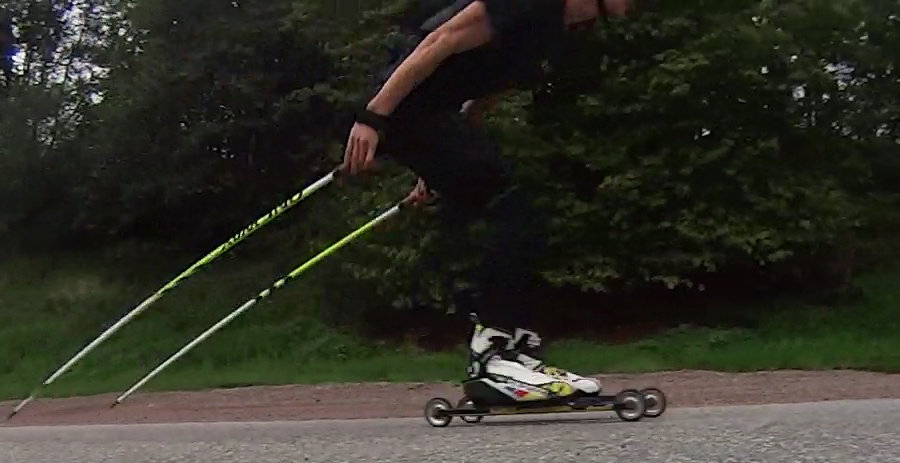}
\includegraphics[width=0.32\textwidth]{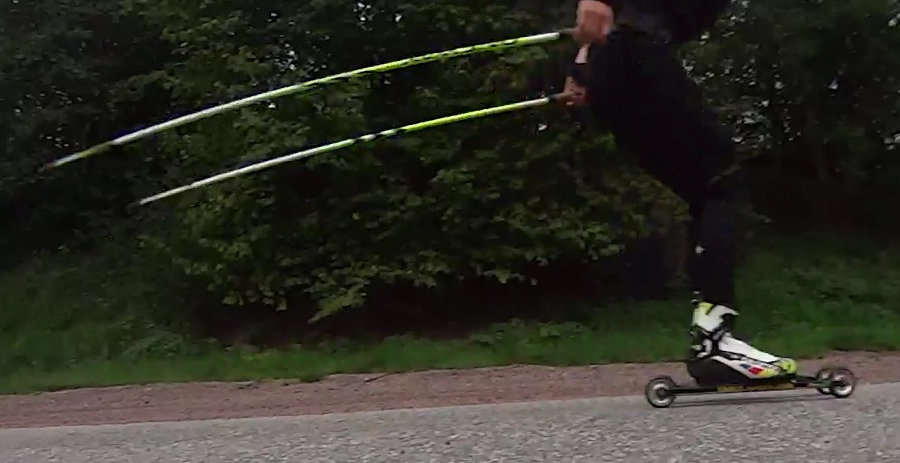}
\caption{Double Poling}
\label{fig:doublePoling}
\end{figure}

\begin{figure}[htbp]
\includegraphics[width=0.32\textwidth]{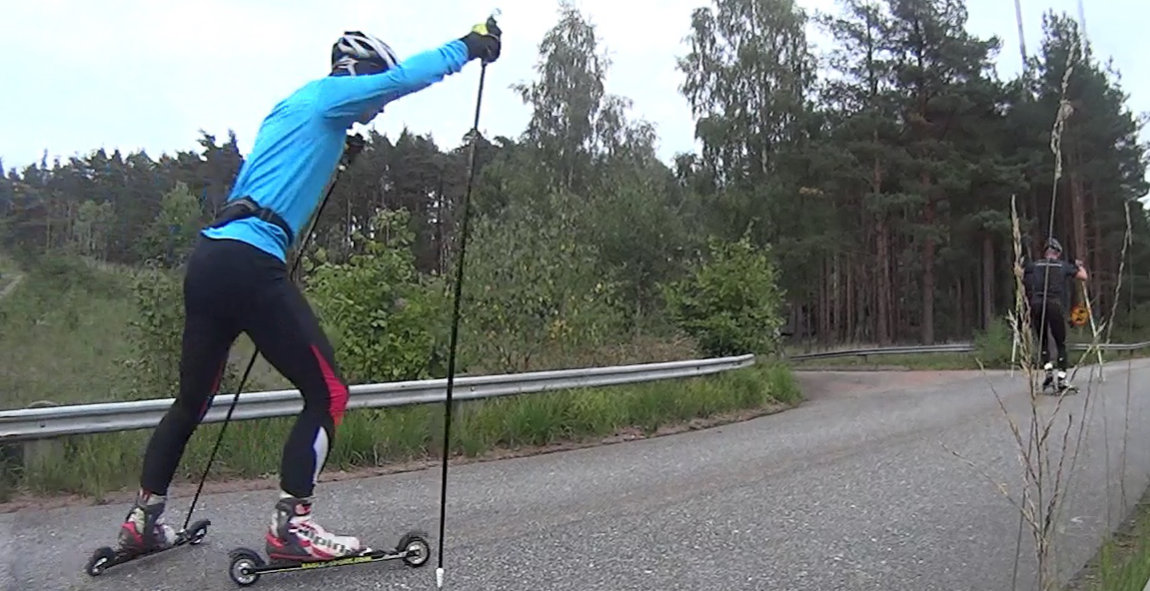}
\includegraphics[width=0.32\textwidth]{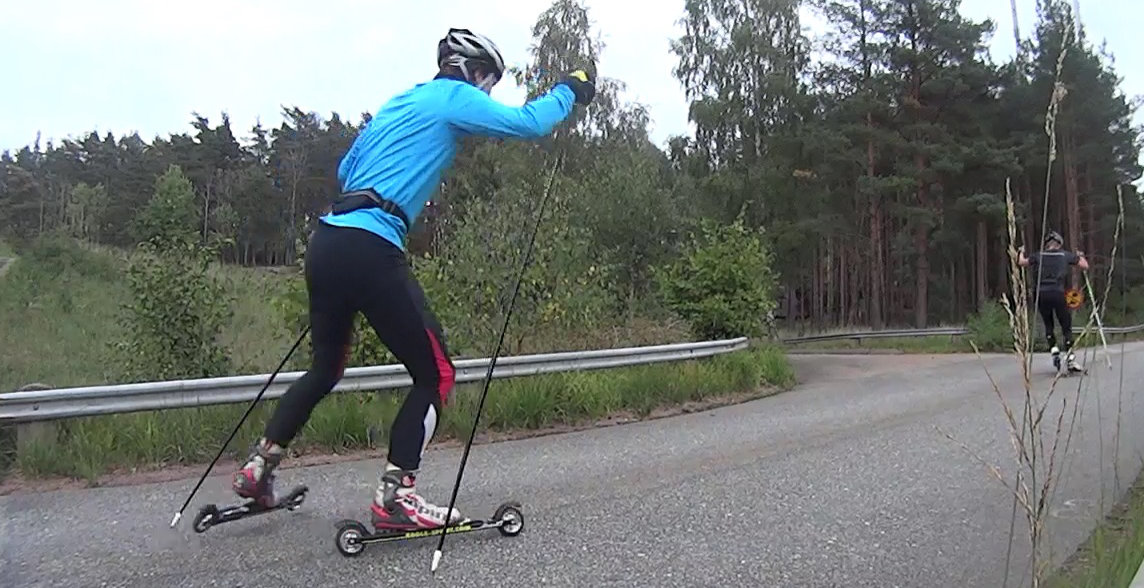}
\includegraphics[width=0.32\textwidth]{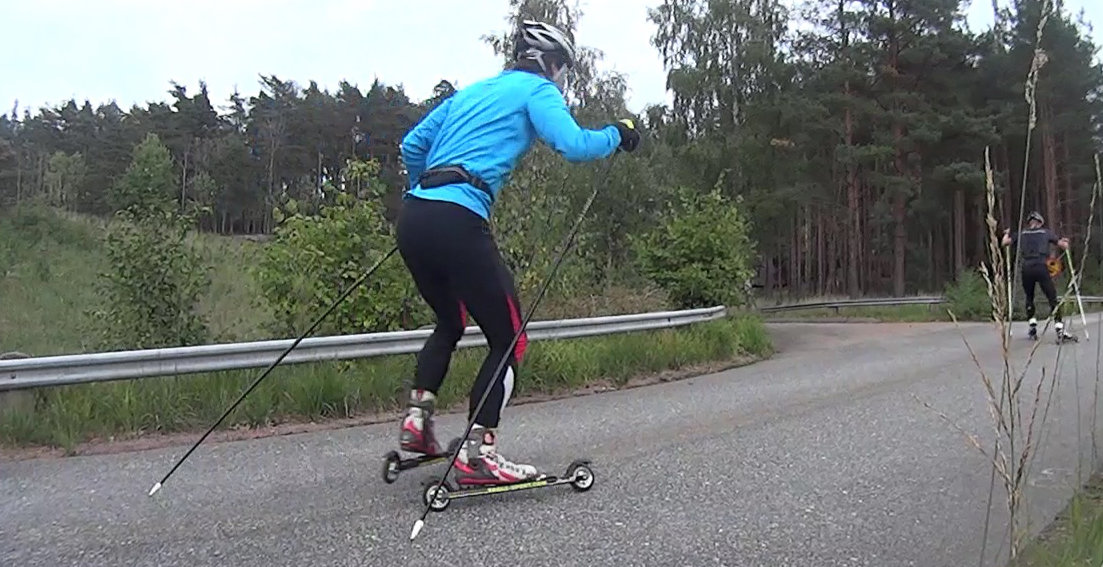}

\includegraphics[width=0.32\textwidth]{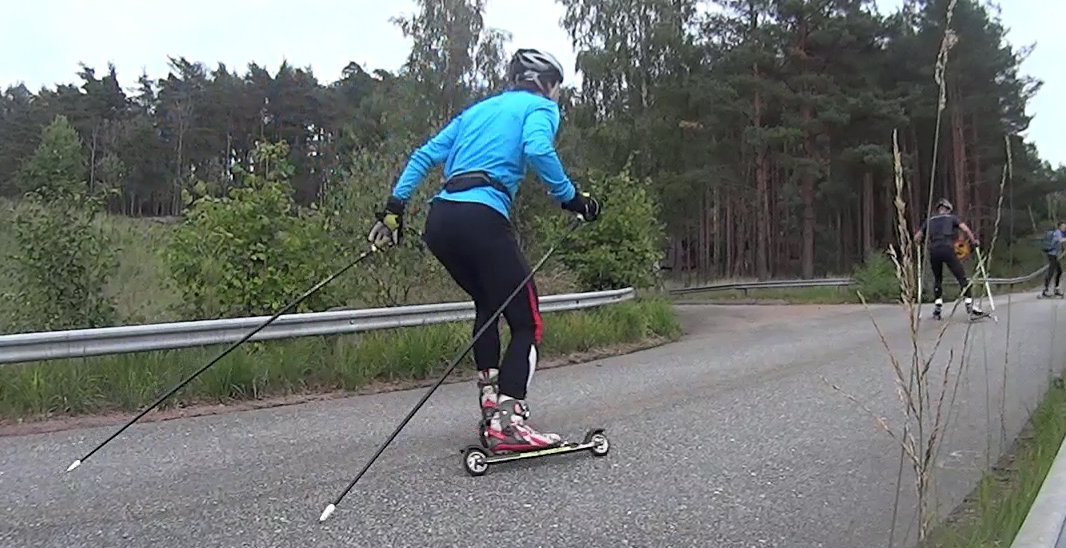}
\includegraphics[width=0.32\textwidth]{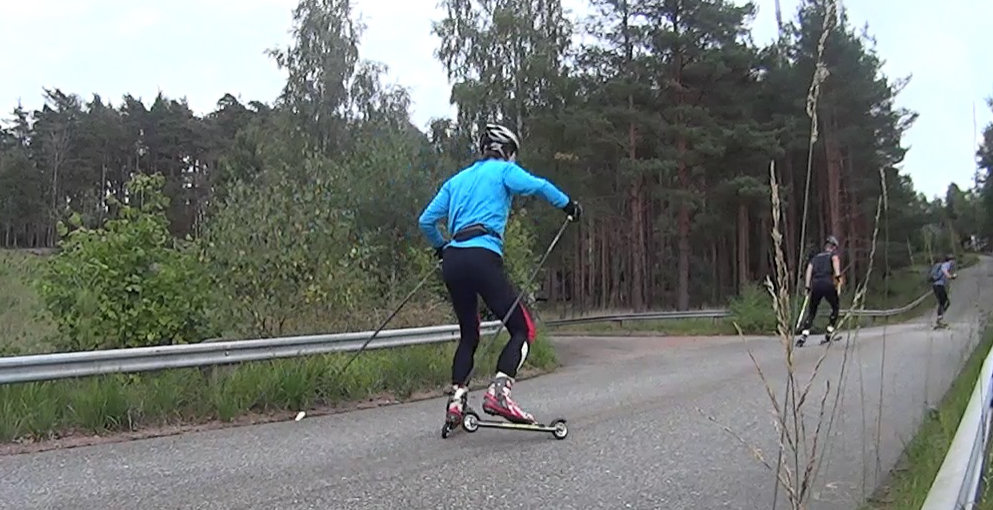}
\includegraphics[width=0.32\textwidth]{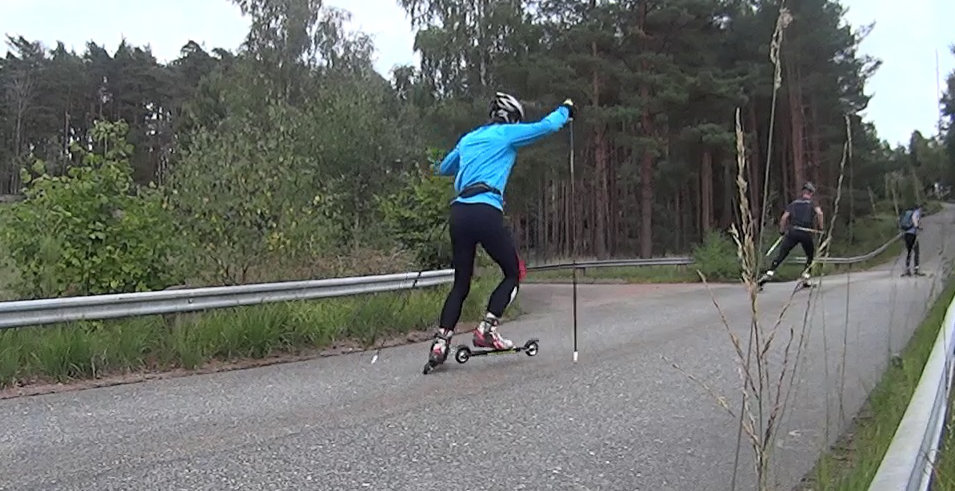}

\caption{Gear 2 skating, leading with right hand.}
\label{fig:gear2}
\end{figure}

\begin{figure}[htbp]
\includegraphics[width=0.32\textwidth]{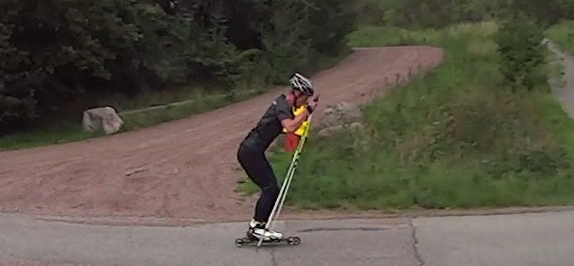}
\includegraphics[width=0.32\textwidth]{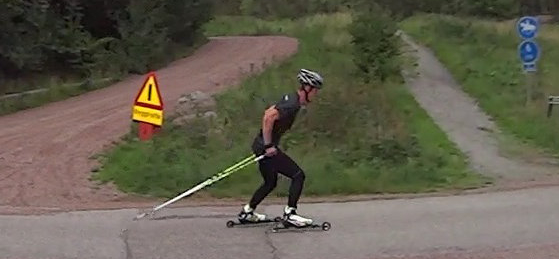}
\includegraphics[width=0.32\textwidth]{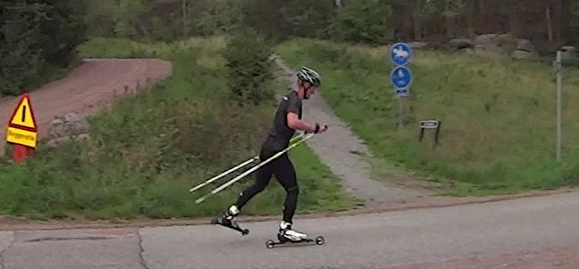}

\includegraphics[width=0.32\textwidth]{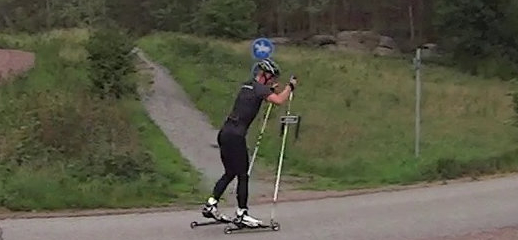}
\includegraphics[width=0.32\textwidth]{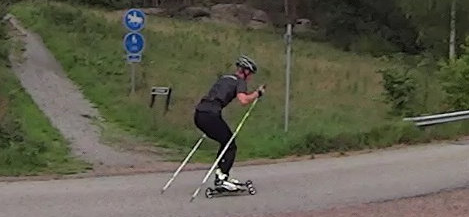}
\includegraphics[width=0.32\textwidth]{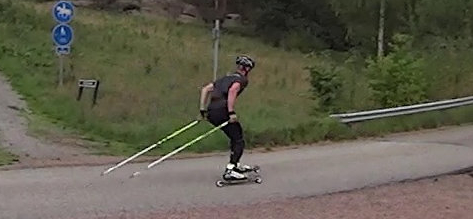}

\caption{Gear 3 skating}
\label{fig:gear3}
\end{figure}

\begin{figure}[htbp]
\includegraphics[width=0.32\textwidth]{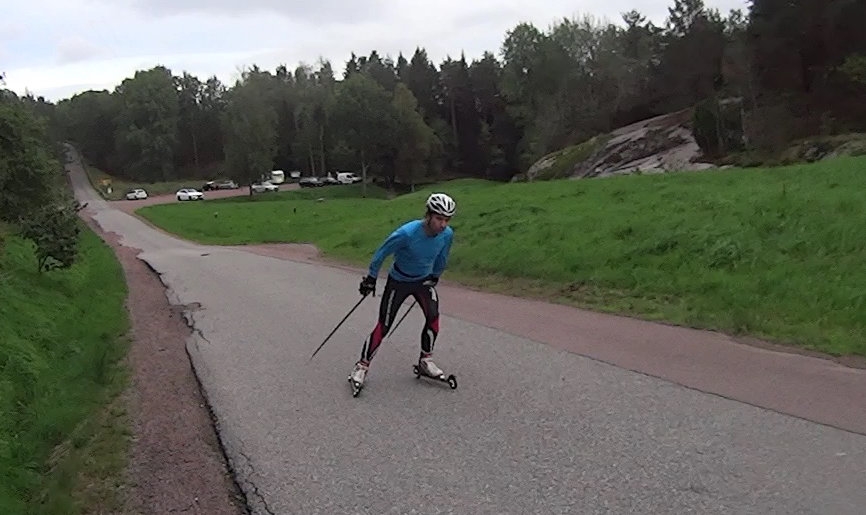}
\includegraphics[width=0.32\textwidth]{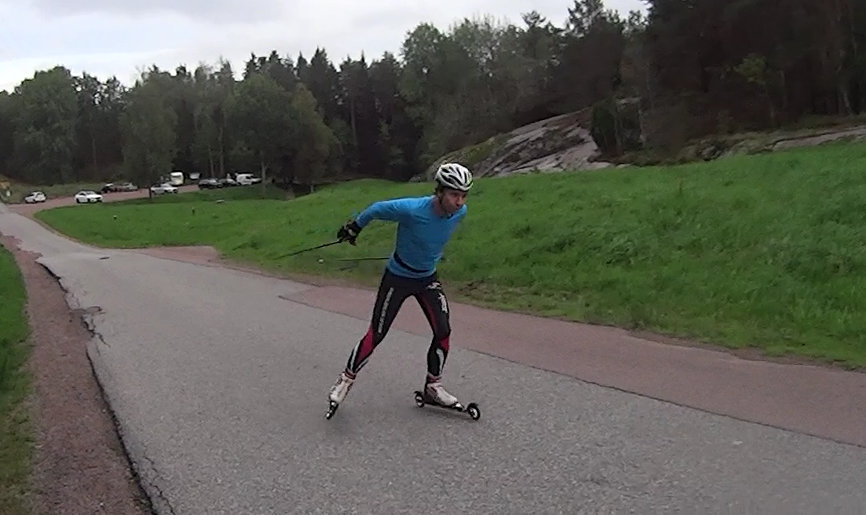}
\includegraphics[width=0.32\textwidth]{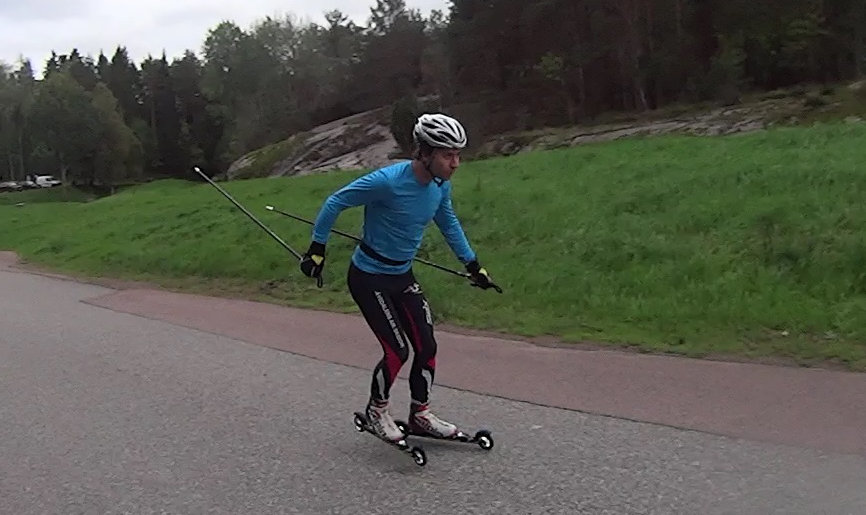}

\includegraphics[width=0.32\textwidth]{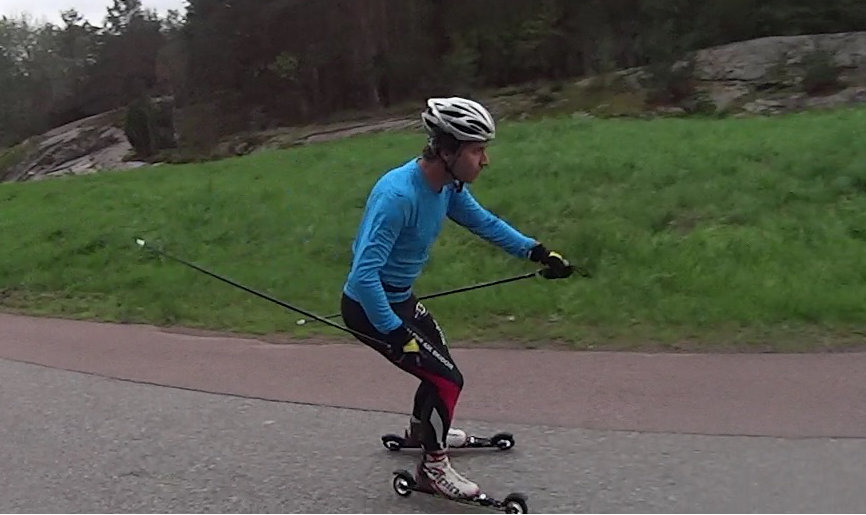}
\includegraphics[width=0.32\textwidth]{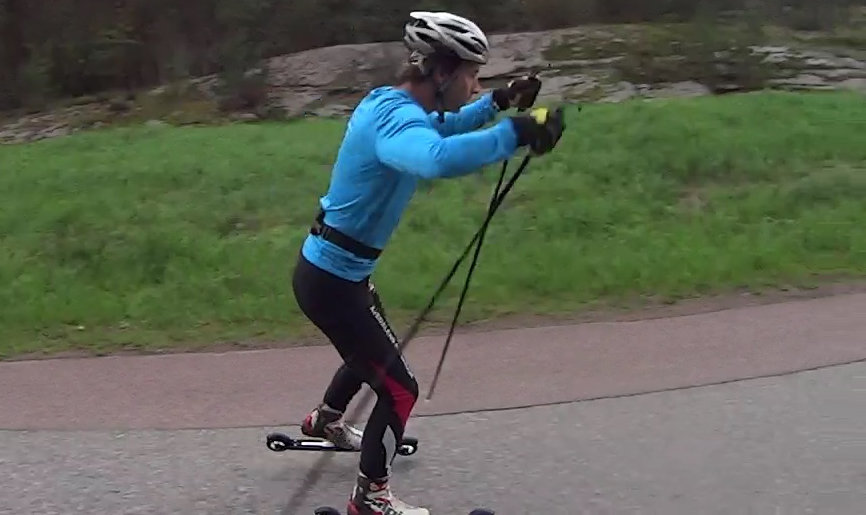}
\includegraphics[width=0.32\textwidth]{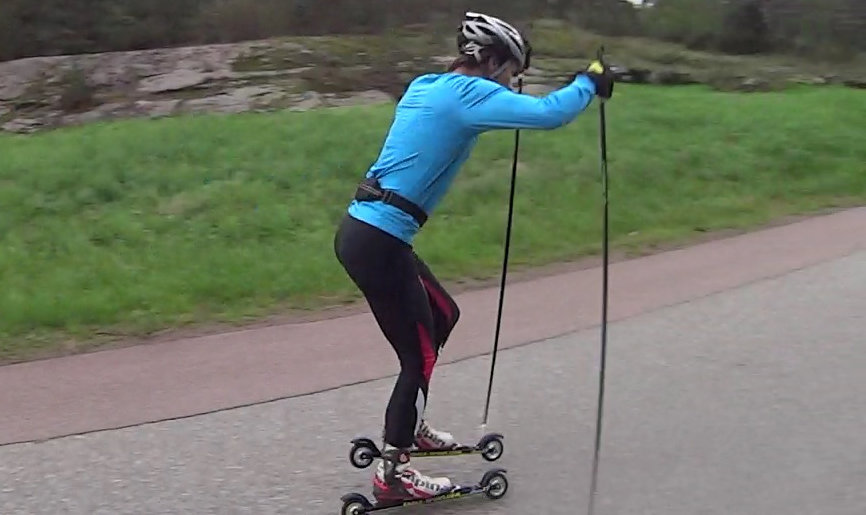}

\caption{Gear 4 skating}
\label{fig:gear4}
\end{figure}

In cross country skiing, several different sub-techniques can be used by the skier, with each technique corresponding to  different motion patterns. The most commonly used skiing techniques are divided into two subgroups -- classical style and freestyle. In this work we focused on four freestyle techniques: double poling (which may also be used in classical style races), and three skating techniques refereed to as Gear 2, Gear 3 and Gear 4 following the notation in \cite{nilsson}\footnote{We note that the notation varies between different countries, these techniques are sometimes also referred to as V1, V2 and V2a. See  \cite{nilsson} for a discussion.}, and illustrated in figures \ref{fig:doublePoling} -- \ref{fig:gear4}.
%\footnote{Readers unfamiliar with cross-country skiing may also view the different techniques in this videoclip: \url{https://youtu.be/AwAGJu\_7xjI}.}. 
There is also a Gear 1, which is rarely used in practice except in extremely steep terrain, and a Gear 5 which only uses the legs and no poling. These styles were not included in this study. 

In double poling, as illustrated in figure \ref{fig:doublePoling}, the skier mostly uses the upper body by moving the arms in parallel. In classical style racing, double poling is the fastest gear, primarily used in horizontal or gentle down-hill terrain, when the velocity is already high, and the skier is not in need of using the legs. In freestyle racing, double poling is not much used, except under special conditions, when there is little space to use the legs in a masstart race or if the snow is very icy, making it difficult to use the legs.

%The technique is mostly used in horizontal terrain, when the velocity is already high, and hence the skier is not in need of using the legs. 

In skating Gear 2 the motion pattern of the skier is asymmetric, with the skier leading with one arm (see figure \ref{fig:gear2}), performing one double pole push for every second leg push. The skier may alternate which arm is leading. Gear 2 is mostly used in uphill or horizontal terrain when the friction is high. Gear 3 is characterised by the skier preforming one double pole push for each leg push (see figure \ref{fig:gear3}). Gear 3 is mostly used in the translation between uphill and downhill, or in horizontal terrain when the skier wants to accelerate to higher speed. The last skating style considered for this work is Gear 4 (see figure \ref{fig:gear4}), which has the same relationship between arms and legs as Gear 2, however the techniques differs in how the poling is preformed: in Gear 4 the skier does the poling symmetrically with respect to each side. Gear 4 is mostly used in horizontal terrain when the snow ski friction is low.
%(From figure \cref{fig:doublePoling} and \cref{fig:Gear234} and by the descriptions of the different gears, 
We note that double poling and Gear 3 have arm-motion patterns that looks considerably like each other. This raised the question of whether separating these two sub-techniques would be more difficult, based on arm movements only.

%For the purpose of this study this is an interesting fact since it raises the question wheather an algorithm have a harder type to separate these two techniques in comparison to the other techniques.

\section{The Dataset}\label{sec:dataset}

The dataset, provided by Skisens AB, consists of data from three individuals (male, experienced recreational skiers) using Skisens ski pole handles with sensors. The data was collected on roller skis on different days, in varied terrain and under varied conditions. There were both uphill and downhill sections as well as turns.
Each skier used the three different skating styles (Gear 2, Gear 3 and Gear 4) plus double poling. For each gear there are a number of disjoint data segments, where each segment is a continuous time-series of data during which the skier only uses a specified style. The data collected is summarised in Table \ref{tab:data}. Data was recorded at 50 Hz (50 samples per second), hence when we refer to time-steps, these are data-points recorded 0.02 seconds apart. After pre-processing the raw data (see section \ref{seq:pre-process}), we extracted a dataset containing 1671 individual strokes\footnote{Of which 252 strokes in Gear 2, 473 in Gear 3, 360 in Gear 4 and 585 strokes using double poling.}.

\begin{table}[]
\centering
\caption {Description of the dataset columns used for machine learning. The coordinate system for the vectors of acceleration and angular velocity is relative to the pole with (a) First axis: Pointing right (orthogonal to pole),
(b) Second axis: Pointing down (parallel to pole), and (c) Third axis: Pointing forward (orthogonal to pole)} 
\begin{tabular}{|l|l|l|l}
\cline{1-3}
\textbf{ No.} & \textbf{Data}    & \textbf{Unit} &  \\ \cline{1-3}
1                      & Time             & second        &  \\ \cline{1-3}
2     & Force in the left pole  & Newton        &  \\ \cline{1-3}
3       & Pole-ground angle of the left pole& degrees        &  \\ \cline{1-3}
4 - 6     & Left angular velocity & rad/s     &  \\ \cline{1-3}
7 - 9        & Left acceleration     & m/s$^2$        &  \\ \cline{1-3}
10     & Force in the right pole        & Newton        &  \\ \cline{1-3}
11    & Pole-ground angle of the right pole  & degree        &  \\ \cline{1-3}
12-14       & Right angular velocity & rad/s     &  \\ \cline{1-3}
15-17      & Right acceleration     & m/s$^2$        &  \\ \cline{1-3}
%10                     & Speed            & metres/second  &  \\ \cline{1-3}
%11                     & Altitude         & metres         &  \\  \cline{1-3}
\end{tabular}
\label{tab:data}
\end{table}

We remark that the data recorded also included the GPS position of the skier, but we choose not to include this information as a feature, as the different techniques naturally had been used at distinct road segments (as some techniques are more natural to use e.g. in uphill terrain. If this was included, the models would end up basing their predictions primarily on GPS-position, ignoring the other features, which would lead to poor performance on unseen data recorded in a different location.

\subsection{Data pre-processing}
\label{seq:pre-process}
To prepare the data set for machine learning, we applied some pre-processing techniques described below. First, the data was smoothed to reduce short-term random variation and irregular noise and split into single strokes.
%\subsection{Smoothing time-series data} 
%Smoothing is a data pre-processing technique used to reduce the short-term, random variations or €œirregular€ noise in the time-series data and provide a more precise prediction of the long-term trend. %A moving average (rolling average or running average or rolling mean) is calculated by generating a series of averages of different subsets of the entire dataset. 
%We resampled the data using the combination of rolling() and mean() functions in the Python library scikit-learn\footnote{\url{http://scikit-learn.github.io/stable}}. The rolling() function which will automatically group the time-series observations into a window, where we can specify the window size. These windows identify sub periods of the time series dataset. Once we have the window, we take the mean value to generate the transformed dataset. 
%Through experimentation, we found that a window size of 3 was suitable for our model.
%\subsection{Splitting to single strokes}
%\label{seq:split_to_strike}
%For this study, we wanted the model to learn classifications for single strokes. 
Secondly, as the data originally formed long time series where different techniques were used, we split them into shorter segments containing one stroke each, with the objective to learn the label for each such segment. Each such one-stroke segment was defined to include the time sequence from the moment when the skier lifts the pole, followed by the next ground contact phase until the skier lifts the pole in the air again. The splitting was implemented by iterating over the entire time series, and splitting, when the force changes from having magnitude larger than a threshold $T$, to smaller than $T$, where $T = 0.4 N$ (motivated by inspection of the data). 
Naturally, not all strokes are of the same length time-wise, hence to make all samples the same length (fitting the input to the classifier) each strike sequence was (if needed) zero padded to have the fixed length of $L = 140$ time steps. As shown in \cref{tab:data}, there are 16 data values recorded for each time-step. Hence, each stroke is represented by a matrix of size 140 x 16.

%\subsection{One-hot encoding of categories}
For efficient implementation of a machine learning algorithm, the categories for the skiing techniques (double poling, gears 2-4) are represented in numerical form, using one-hot encoding, where a new binary variable is added for each of the four category. %A 1€ value is placed in the binary variable for the gear and a 0€ values for the other gears, as shown in \cref{tab:one-hot-encoding}, resulting in a vector representing each category.

%\begin{table}[htb]
%\centering
%\caption {One-hot encoding applied to the gear categories} 
%
%\begin{tabular}{llll}
%\hline
%\multicolumn{1}{|l|}{\textbf{Gear 2}} & \multicolumn{1}{l|}{\textbf{Gear 3}} & \multicolumn{1}{l|}{\textbf{Gear 4}} & \multicolumn{1}{l|}{\textbf{Double Poling}} \\ \hline
%\multicolumn{1}{|l|}{1}              & \multicolumn{1}{l|}{0}              & \multicolumn{1}{l|}{0}              & \multicolumn{1}{l|}{0}                     \\ \hline
%\multicolumn{1}{|l|}{0}              & \multicolumn{1}{l|}{1}              & \multicolumn{1}{l|}{0}              & \multicolumn{1}{l|}{0}                     \\ \hline
%\multicolumn{1}{|l|}{0}              & \multicolumn{1}{l|}{0}              & \multicolumn{1}{l|}{1}              & \multicolumn{1}{l|}{0}                     \\ \hline
%\multicolumn{1}{|l|}{0}              & \multicolumn{1}{l|}{0}              & \multicolumn{1}{l|}{0}              & \multicolumn{1}{l|}{1}                     \\ \hline
%\end{tabular}
%\label{tab:one-hot-encoding}
%\end{table}

\section{Machine learning models} \label{sec:models}
We experimented with three different types of deep machine learning models for stroke classification: a long short term memory network (LSTM) \cite{LSTM}, a bidirectional long short term memory network (BLSTM) \cite{BLSTM}, and a one dimensional convolutional neural network (CNN). The models were implemented in Python using the Keras/TensorFlow libraries\footnote{\url{https://www.tensorflow.org/guide/keras}}. The code is available online\footnote{\url{https://github.com/moajohansson/ai-in-sports}}.

\subsection{Long short-term Memory (LSTM)}
\label{seq:LSTM_Moel_1}
An LSTM network \cite{LSTM} is a type of recurrent neural network, which, unlike for instance CNNs, is able to pass some information along from previous steps in e.g. a time sequence. LSTM's contain special memory gates which enable some long-term dependencies to also be captured by the network during training, addressing a weakness of standard recurrent neural networks which might suffer from vanishing error gradients during training. LSTMs are suitable for time series data, and have successfully been used in for example many natural language tasks.

\begin{figure}[htbp]
    \centering
    \captionsetup{width=0.8\textwidth}
    \includegraphics[width=8cm]{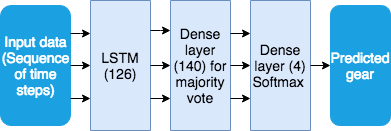}
    \caption{Network architecture for the LSTM model, with an LSTM cell with two dense layer. The light blue boxes indicates layers in the network, and the number of neurons in each layer is stated inside the brackets in each layer.}
    \label{fig:Lstm_Strikes_Model}
\end{figure}

The LSTM model in our experiment combines an LSTM cell with two dense layers (see \cref{fig:Lstm_Strikes_Model}). The input of the LSTM model is a sequence of $140$ data points, each corresponding to one pole push. % as discussed in section \cref{seq:split_to_strike}.  
The first layer of the LSTM model is an LSTM cell with 126 neurons, chosen experimentally from the set $[26,64,126,256]$ for minimising the error on the validation set.  
The second layer of the model is a dense layer with 140 neurons, which is connected to a dense layer with 4 neurons and a softmax activation function. These two layer's can be interpreted as a weighted majority vote, it weights the importance of each 140 time steps and then gives one result of the most likely gear for the entire pole push. 
Besides using a layer for majority voting, as in the model above, we also examined the performance when performing majority voting after the model had classified each of the time steps separately in the pole push. However, employing weighted majority voting as layer in the model improved the accuracy on validation data with almost $10\%$, in comparison to performing majority voting after classifying each time step. %It hence seems like some of the samples in the sequence are more important when predicting the gear for the entire strike. 

\subsection{Bi-directional LSTM Model}
\label{seq:BLSTM_model}
The BLSTM network \cite{BLSTM}, has similar network architecture as the LSTM network. The difference is that the LSTM network passes information only in the forward direction, whereas the BLSTM network passes information in both the forward and backward direction. Hence, a BLSTM cell specified with same number of neurons as an LSTM cell, but uses twice as many  weights. 

\begin{figure}[ht]
    \centering
    \captionsetup{width=0.8\textwidth}
    \includegraphics[width=8cm]{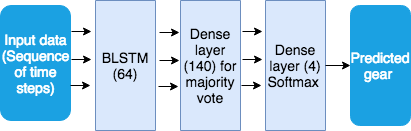}
    \caption{The network architecture for the BLSTM model with one BLSTM layer and two dense layers.}
    \label{fig:BLSTM_Model}
\end{figure}
Our BLSTM model consist of one BLSTM cell and two dense layers, see \cref{fig:BLSTM_Model}. 
Experimentally minimising validation set error suggested setting the number of neurons in the BLSTM cell to 64. Further, the number of neurons in the two dense layers was chosen to be 140 and 4 respectively, as in the LSTM model. 

\subsection{Convolutional Neural Network Model}
\label{seq:CNN_Model}
\begin{figure}[htbp]
    \centering
    \captionsetup{width=0.8\textwidth}
    \includegraphics[width=8cm]{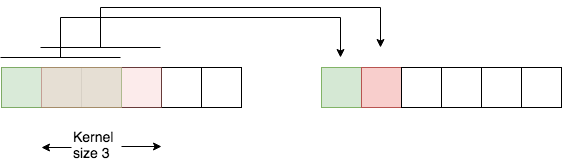}
    \caption{The convolutional operator in a one dimensional CNN network, with kernel size 3 and stride 1.}
    \label{fig:conv_operator}
\end{figure}

CNN are a deep neural network architecture which has primarily been used for image processing. The CNN network employs a convolutional operator which performs a kind of down-sampling, as illustrated in \cref{fig:conv_operator}. For image processing, two dimensional CNNs are typically used, but as we here deal with time-series, we employ a one-dimensional CNN acting in the time-dimension. As seen in \cref{fig:conv_operator}, the \emph{kernel size} determines how many of the input elements will be weighted and summed together in each convolutional operation, while the \emph{stride} determines how many steps to move the kernel for each operation.% In \cref{fig:conv_operator} the stride is 1, and hence the kernel is moved one step to the left for each operation. 

\begin{figure}[htbp]
    \centering
    \captionsetup{width=0.8\textwidth}
    \includegraphics[width=12cm]{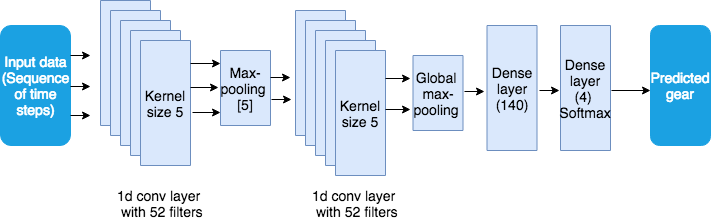}
    \caption{The network architecture for the CNN model. }
    \label{fig:CNN_Model}
\end{figure}

%Explain the architecture of the best model + a table giving all the combinations of parameters
Our CNN model consists of two one dimensional convolutional layers and two dense layers (see \cref{fig:CNN_Model}), as well as max-pooling and global max-pooling layers. The latter two layers  are used for down-sampling, locally and globally.

Based on experimental evaluation minimising error on the validation data we choose $52$ filters in each convolution layer. The model performance using one convolutional layer was also tested, but the model using two convolutional layers performed better on validation data. Similarly, the kernel size was set to $k = 5$, and the pool-size in the max-pooling layer was also set to 5. The number of neurons in the two dense layers was chosen to be 140 and 4 respectively, as in the LSTM model. 

\section{Experiments and Results} \label{sec:experiments}
In this section we present classification results for the three models (LSTM, CNN, BLSTM) described above. The experiments were run on a Macbook Air with an Intel Core i5 1,7 GHz processor and 4GB of memory. 

\subsubsection*{Experiment 1: }
We trained the models on a subset of the data containing samples from all three skiers, and evaluated on another unseen subset as test data. We suspect that the same person performs strokes in the same techniques in a relatively consistent manner, hence the strokes in the test set are likely to  be quite similar to something from the training set. A motivation for this kind of experiment is envisaging an application using Skisense-sensors which is personalised to the owner, who initially ``calibrates" the product by skiing in specified gears to collect personal training data.  

Experiment 1 was performed for all three models described above, using five-fold cross-validation, with each fold containing approximately the same number of strokes and the same proportion of strokes in each gear (folds 1-4 of 329 strokes, fold 5 of 355 strokes, from the total dataset of 1671 strokes).  
\begin{table}[htbp]
\centering
\caption{Accuracy results for experiment 1, using five fold cross-validation}

\begin{tabular}{l  | r}
\textbf{Model}  & Accuracy    \\ \hline
LSTM                      & 0.95        \\ \hline
CNN                         & 0.90        \\ \hline
BLSTM                     & 0.95        
\end{tabular}
%\captionsetup{width=0.8\textwidth}
\label{tab:res1}
\end{table}

The results are promising, with between 90-95\% correct classifications on average over the five folds, as summarised in Table \ref{tab:res1}. We note that the CNN model performed slightly worse than the other two, and also that the performance differed more over the different folds for the CNN model. We suspect that the CNN model suffered more than the LSTM-based models from the relatively small dataset. We note that the LSTM-based models also contains more trainable parameters than the CNN-model, so more experimentation is needed with different CNN architectures. Training takes longer for the LSTM and BLSTM models, approximately 1-2 hours on the laptop computer used, compared to around 10 minutes for the CNN model. We note that for a larger study, we would use modern hardware which would considerably speed up training.

\begin{figure}[ht]
    \centering
    \captionsetup{width=0.8\textwidth}
    \includegraphics[width=9cm]{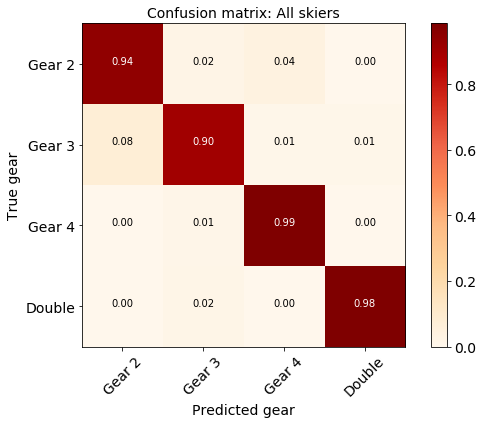}
    \caption{Confusion matrix for LSTM model, experiment 1.}
    \label{fig:LSTM_1_Conf}
\end{figure}

In \cref{fig:LSTM_1_Conf} the confusion matrix for the LSTM model is presented (the other two models had very similar results). We note that Gear 4 and double poling were the easiest to classify, while Gear 3 was the hardest. This was somewhat surprising, as the arm movements of gear 4 and double poling are visually quite similar. 
%We note from \cref{fig:LSTM_1_Conf}, that the model seems to have almost equal capacity of separating the different gears. %However, the model seems to be slightly better in classifying data from the double poling, this might be caused by the fact that the skiers use in general more force when double poling, which make it easier to distinguish from the other gears.  

\subsubsection*{Experiment 2: }
Experiment 1 does not test the models' capability to generalise to a person it has not seen before. This was somewhat difficult to test, due to the small dataset. However, we did a second experiment with the best-performing model from Experiment 1 (the LSTM model) where we trained on data from two skiers, and evaluated on unseen data from the third individual. This was expected to be harder, as the model would have to generalise, and ideally learn how an "average" stroke in each technique would be represented by the sensor data.
As expected, performance dropped to 78\%. We believe that this could be improved by training on a larger dataset with samples from many individuals, and performing a larger study is future work.

\section{Discussion and Further Work}
We have conducted a pilot study using data from sensors fitted to ski pole handles to predict which technique or gear the skier is using. The pilot experiment aimed at classifying time-series for single strokes, as these are easy to identify from the power data recorded from the poles (near-zero readings indicating when the poles are in the air). We have not yet attempted the task of passing in continuous sequences of skiing strokes and identifying gear changes. This is an interesting problem, as some previous work, e.g. \cite{Rindal2017}, report that mis-classifications of single strokes often happen near change points. 

For this study we only had access to data from three individuals, resulting in a dataset of merely 1671 strokes, which is on the small side for deep learning. This was noticeable in Experiment 2, where, unsurprisingly, classification accuracy dropped when the model was presented with an unseen skier. We are however encouraged by the results in this study to gathering a larger dataset and performing a larger evaluation in the near future. 
Most other works in cross-country skiing technique classification come from the sports science domain, and often include only a few individuals in the studies (e.g. 10 skiers in \cite{Rindal2017}, four skiers in \cite{Jang2018}). Furthermore, these studies often primarily focus on reaching high accuracy for these specific individuals (often elite athletes). Experiments are often in the style of our Experiment 1, i.e. the training data and test data contain the same individuals. As future work, it would be very interesting to apply deep learning techniques to a much larger dataset, containing both professionals and recreational skiers and investigate whether one can train a model to generalise well enough on all individuals, without taking small individual variations into account. This is particularly relevant from the perspective of Skisens, as they are interested in including technique classification together with their ski-pole sensors in for example a smart sports watch. Ideally, one would like to have a pre-trained model which does an acceptable job out of the box, and possibly then adapts to the individual user, without having to be trained from scratch. %This is called \emph{transfer learning} and has successfully been employed in e.g. natural language processing and object recognition in images \cite{transfer}.

\bibliographystyle{abbrv}
\bibliography{bibfile}

\end{document}